\documentclass[preprint,12pt]{elsarticle}

\usepackage{epsfig}

\usepackage{amssymb}
\usepackage{amsmath,amssymb,amsfonts}
\usepackage{ragged2e} 
\usepackage{booktabs,makecell, multirow, tabularx}
\usepackage{amsmath}
\usepackage{autobreak}

\journal{Medical Image Analysis}

\begin{document}

\begin{frontmatter}



\title{A Labeled Ophthalmic Ultrasound Dataset with Medical Report Generation Based on Cross-modal Deep Learning}

\author{Jing Wang\fnref{label1}}
\ead{jwang@ncut.edu.cn}
\author{Junyan Fan\fnref{label1}}
\ead{fanjunyan218@mail.ncut.edu.cn}
\author{Meng Zhou\fnref{label1}}
\ead{zhoumeng@ncut.edu.cn}
\author{Yanzhu Zhang\fnref{label2}\corref{cor1}}
\ead{syzd710471@163.com}
\author{Mingyu Shi\fnref{label3}}
\ead{myshi@cmu.edu.cn}
\cortext[cor1]{Corresponding author}

\affiliation[label1]{organization={School of Electrical and Control Engineering},
            addressline={North China University of Technology}, 
            city={Beijing},
            postcode={100144}, 
            country={China}}
        
\affiliation[label2]{organization={School of Automation and Electrical Engineering},
        	             addressline={Shenyang Ligong University},
        	             city={Shenyang},
        	             postcode={110159},
        	             country={China}}
        
\affiliation[label3]{organization={Department of Ophthalmology},
        	             addressline={The Fourth Affiliated Hospital of China Medical University},
        	             city={Shenyang},
        	             postcode={110005},
        	             country={China}}

\begin{abstract}
Ultrasound imaging reveals eye morphology and aids in diagnosing and treating eye diseases. However, interpreting diagnostic reports requires specialized physicians. We present a labeled ophthalmic dataset for the precise analysis and the automated exploration of medical images along with their associated reports. It collects three modal data, including the ultrasound images, blood flow information and examination reports from 2,417 patients at an ophthalmology hospital in Shenyang, China, during the year 2018, in which the patient information is de-identified for privacy protection. To the best of our knowledge, it is the only ophthalmic dataset that contains the three modal information simultaneously. It incrementally consists of 4,858 images with the corresponding free-text reports, which describe 15 typical imaging findings of intraocular diseases and the corresponding anatomical locations. Each image shows three kinds of blood flow indices at three specific arteries, i.e., nine parameter values to describe the spectral characteristics of blood flow distribution. The reports were written by ophthalmologists during the clinical care. The proposed dataset is applied to generate medical report based on the cross-modal deep learning model. The experimental results demonstrate that our dataset is suitable for training supervised models concerning cross-modal medical data.
\end{abstract}

%

\begin{keyword}
 Medical report generation\sep Ophthalmic ultrasound\sep Computer vision\sep Natural language processing
\end{keyword}

\end{frontmatter}


\section{Introduction}
Medical imaging plays a crucial role on the clinical diagnosis and treatment especially in the field of ophthalmology. The popular imaging techniques include fundus photography, optical coherence tomography (OCT), and fluorescein angiography of the retina. The adequate interpretation of the ophthalmic examination requires perfossional ophthalmologist or radiologists. Due to the increasing workloads, ophthalmologists face significant time and effort limitations in analyzing medical images and generating diagnostic reports. Moreover, the subjective expertise of different ophthalmologist will result the potential variations in the interpretation of the same image. The existing medical report generation has mostly focused on the radiographic images, particularly chest X-ray images. Various medical report generation datasets have been released for different medical modalities, such as fluorescein angiography (FFA) images \cite{ref1}, lung CT scans \cite{ref2}, and color fundus photography (CFP) \cite{ref3}. There is a lack of research specifically on ophthalmic ultrasound images and report generation. Moreover, most of the existing datasets are in English. There is a significant research gap in the field of Chinese medical report generation. Therefore, the development of annotated ophthalmic ultrasound image datasets with corresponding report is necessary for the ophthalmic diagnosis based on artificial intelligence (AI).

Inspired by the image captioning task, an increasing number of researchers have applied this approach to the generation of medical reports. Its primary objective is to provide annotation for subsequent diagnosis and treatment of diseases. Many deep learning-based automatic generation methods have been proposed to lighten the workload of doctors \cite{ref4,ref5,ref6,ref7,ref8}. These methods usually use the convolutional neural network (CNN) \cite{ref9,ref10} to extract visual feature,  and use the recurrent neural network (RNN) to predict reports \cite{ref11}. They have given a preliminary feasible scheme for medical report generation (MRG). Latter text-image attention mapping is explored to explain the automatic generation process although its accuracy is not yet known \cite{ref12}. There still exist many challenges which limits the practical application of deep learning method as follows. (1) The construction of large and specific medical dataset is time-consuming and labor-intensive; (2) Due to the complexity of medical image interpretation, The accuracy achieved by deep learning model is not up to the level of specialized doctors.

In this paper, we propose a labeled ophthalmic dataset of ocular ultrasound image, text report and blood flow information. To the best of our knowledge, it is the only ophthalmic dataset that contains all the three modal information simultaneously. A comprehensive experiment is conducted on this dataset, and the result demonstrates that the proposed dataset is suitable for training all kinds of supervised learning models which concerns the cross-modal medical data. The main motivation for the dataset construction is as follows. From the view of scientific research, it is helpful to develop medical AI learning algorithm fusion the computer vision and natural language processing. It can be applied but not limited to the cross-modal report generation. From the view of clinical applications, this dataset provides the ultrasound images which give insight into the morphology and structure of eyes. It includes 15 common ophthalmic diseases, such as retinal detachment, choroidal detachment, vitreous stellate degeneration, vitreous hemorrhage, endophthalmitis and vitreous opacity. The rich types disease is to broaden the learning model for the initial diagnosis and treatment.

The main contributions of this paper are summarized as follows:
\begin{itemize}
	
	\item A large-scale medical dataset is constructed which is comprised of 4,858 eye ultrasound images and their corresponding Chinese reports. All the data were collected from the real-world clinical practices, and the report accurately represent the writing patterns of ophthalmologists. The dataset facilitates cross-modal learning and report generation whose text pattern is more aligned with clinical practices.
	
	\item Compared to other datasets, the proposed dataset includes additional blood flow parameter information extracted from the ultrasound examination beside the image and text. These parameters describe the spectral characteristics of blood flow distribution at three specific arteries. It plays a key role on assisting medical diagnosis and treatment decisions.
	
	\item Cross-modal memory network is given to generate report according to the proposed dataset. A comprehensive experiment is conducted, and the predict accuracy of medical report is evaluated based on the NLG metrics. The result shows that our dataset can be applied to the medical report generation. It is helpful to drive the AI based ophthalmic medical diagnosis.
	
\end{itemize}

The rest of the paper is organized as follows: Section 2 reviews various existed medical datasets and medical report generation (MRG) methods. Section 3 presents the construction methodology of the labeled ophthalmic ultrasound dataset. Section 4 gives a comprehensive report generation experiment based on the cross-modal memory network. Section 5 draws a conclusion as well as future perspective.

\section{Related Work}
Many kinds of medical image have been widely utilized to develope an AI-aided diagnosis system. Here we reviewed the existed medical image datasets and MRG methods.

\subsection{Medical report datasets}
Medical report generation has received an increasing attention in the fields of AI aided clinical medicine. Many medical report datasets have been proposed, such as Open-IU \cite{ref13}, DEN \cite{ref3}, RDIF \cite{ref33} and COV-CHR \cite{ref2}. We compared our ocular ultrasound dataset with 14 datasets used for medical report generation. The relevant statistics are given in Table 1. Open-IU and MIMIC-CXR \cite{ref14} are two widely used benchmarks for medical report generation. The Open-IU Indiana University chest X-ray dataset contains 8,121 images associated with 3,996 de-identified radiology reports. MIMIC-CXR is the largest public X-ray dataset, containing 473,057 chest X-ray images and 206,563 reports. It also provides more relevant disease impressions that can be used for disease classification. The PEIR Gross \cite{ref12} dataset consists of 7,442 pairs of images from 21 different subcategories. Unlike Open-IU, each image in PEIR Gross dataset is associated with only one sentence in the corresponding report. PADCHEST \cite{ref15} is a labeled large-scale, high resolution chest X-ray dataset for the automated exploration of medical images along with their associated reports. Fetal ultrasound dataset \cite{ref31} consists of 2,800 frames extracted from videos, along with the corresponding reports. Due to the characteristics of the Chinese vocabulary, the average report lengths of CX-CHR \cite{ref16} and COV-CHR \cite{ref2} are much larger than those of the English medical report dataset.

There are also five retinal datasets, including retinal images and text. FFA-IR \cite{ref1} provides interpretable annotations by labeling 46 foci in a total of 12,166 regions, as well as the Fundus Fluorescein Angiography(FFA) images and reports. It plays an important role in identifying the disease and generating report. In contrast to FFA-IR, DEN primarily consists of color fundus photography (CFP) images (13,898 CFP and 1,811 FFA). STARE \cite{ref17} released a total of 397 images including CFP and FFA in 2004. However, the text provided with these images is short free-text diagnostic labels, rather than observational reports of image findings. It is therefore not suitable for training the medical report generation model. DIARETDB1 \cite{ref18} has good annotation of lesion location and size, but the number of CFP images is limited. MESSIDOR \cite{ref19} includes 1200 CFP images and 600 Fine-Great French reports. In summary, it can be seen that most of the current datasets are for chest diseases. Not much research has been done on datasets for eye diseases. The existed datasets for eyes are basically color fundus images, which are generally used for screening, diagnosis and monitoring of fundus diseases, such as retinopathy and macular degeneration. However, it is difficult to assess some intraocular diseases such as vitreous humor and retinal detachment. Unlike the existing medical reports, our dataset builds image report pairs for clinically collected ocular ultrasound images, Chinese reports and additional blood flow parameter information. It will play an important role in the diagnosis of ocular diseases and automated report generation studies.


\begin{table*}	
	\begin{center}
		\caption{Summary of medical datasets}
		\scalebox{0.7}{
			\label{tab:table1}
			\begin{tabular}{ccccc}		
				\hline \textbf{Name of Dataset} & \textbf{Image Modality} &  \textbf{Number of Images}& \textbf{Report Cases} &  \textbf{Report Language} \\			
				\hline RDIF\cite{ref33} & Kidney Biopsy & 1152 & 144 & English \\
				COV-CHR\cite{ref2} & Lung CT-Scans & 728 & 728 & English/Chinese \\				
				Fetal Ultrasound\cite{ref31} & Fetal ultrasound & 2800 & 2800 & English \\
				PEIR Gross\cite{ref12} & Gross lesions & 7,442 & 7,442 & English \\
				Open-IU\cite{ref13} & Chest X-Ray & 7470 & 2955 & English \\
				MIMIC-CXR\cite{ref14} & Chest X-Ray & 377,110 & 276,778 & English \\
				PADCHEST\cite{ref15} & Chest X-Ray & 160,868 & 22,710 & Spanish \\
				CX-CHR\cite{ref16} & Chest X-Ray & 45,598 & 40,410 & Chinese \\
				TJU\cite{ref34} & Chest X-ray & 19,985 & 19,985 & Chinese \\						
				STARE\cite{ref17} & CFP+FFA & 397 & 397 & English \\
				DEN\cite{ref3} & CFP+FFA & 15709 & 15709(Keywords) & English \\
				DIARETDB1\cite{ref18} & CFP & 89 & 89 & English \\
				MESSIDOR\cite{ref19} & CFP & 1200 & 587 & French \\				
				FFA-IR\cite{ref1} & FFA & $1,048,584$ & 10790 & English/Chinese \\
				\hline Our & Ocular ultrasound & 4858 & 4858 & Chinese \\
				\hline
		 \end{tabular}}
	\end{center}
\end{table*}

\subsection{Medical Report Generation Model}
Researchers have utilized medical image report datasets to develop medical report generation(MRG) methods. Jing et al. \cite{ref12} proposed an encoder-decoder framework with a co-attention mechanism to exact the visual and textual feature, which simultaneously predicted a medical tag and generated a single sentence. Xue et al. \cite{ref6} generated multiple sentences by fusing multiple image modalities using topic-level LSTM and word-level LSTM \cite{ref11}. Li et al. \cite{ref16} summarized a template library and jointed the retrieval and generation operations through reinforcement learning to select sentences directly from the template library. The medical knowledge graph is embedded into the recursive generative network to improve the learning accuracy \cite{ref20,ref21}. Li et al. \cite{ref2} replaced the multi-level recurrent network with a medical tag graph Transformer to process the long sequence in medical tagging. Chen et al. \cite{ref23} proposed a memory-driven Transformer model to improve the memory capacity during the decoding procedure for the retention and utilization of relevant information. Zhang et al. \cite{ref24} improved the attention mechanism to garantee the model focus on the correct region of medical image, and it also provided interpretable analysis for the diagnostic process. Wang et al. \cite{ref5} proposed a textual image embedding model that generated medical report using the attention mechanism. The model incorporated an attention distribution graph and text embedding information to improve the classification accuracy. Zeng et al. \cite{ref25} generated a ultrasound image report using the target detection algorithm. It detected the lesion region, and generated a medical diagnostic report by encoding and decoding the ultrasound image vector.

\section{Labeled Ophthalmic Dataset Construction}
The dataset included three data modalities: image, free-text report and blood flow parameter data. Different modal data were extracted independently from the same report, then composed into database. Patient information was removed to protect privacy.

The framework for dataset construction is shown in Figure 1. It consists of image cropping, report extraction and blood flow information extraction, which are explained latter. Additionally, this framework gives a foundational illustration about how to apply this dataset for cross-modal report generation. During the report generation, visual features are extracted from the ultrasound images by a traditional CNN \cite{ref10}, then injected into the image encoder. Then the encoded image features are fed into a sentence decoder with embedded text features to generate reports. Usually the sentence decoder is designed as LSTM network \cite{ref11} or Transformer decoder \cite{ref22}, in which cross-modal attention is utilized to align image and text information.

\begin{figure}[htbp] 
	\centering 
	\includegraphics[width=1\textwidth]{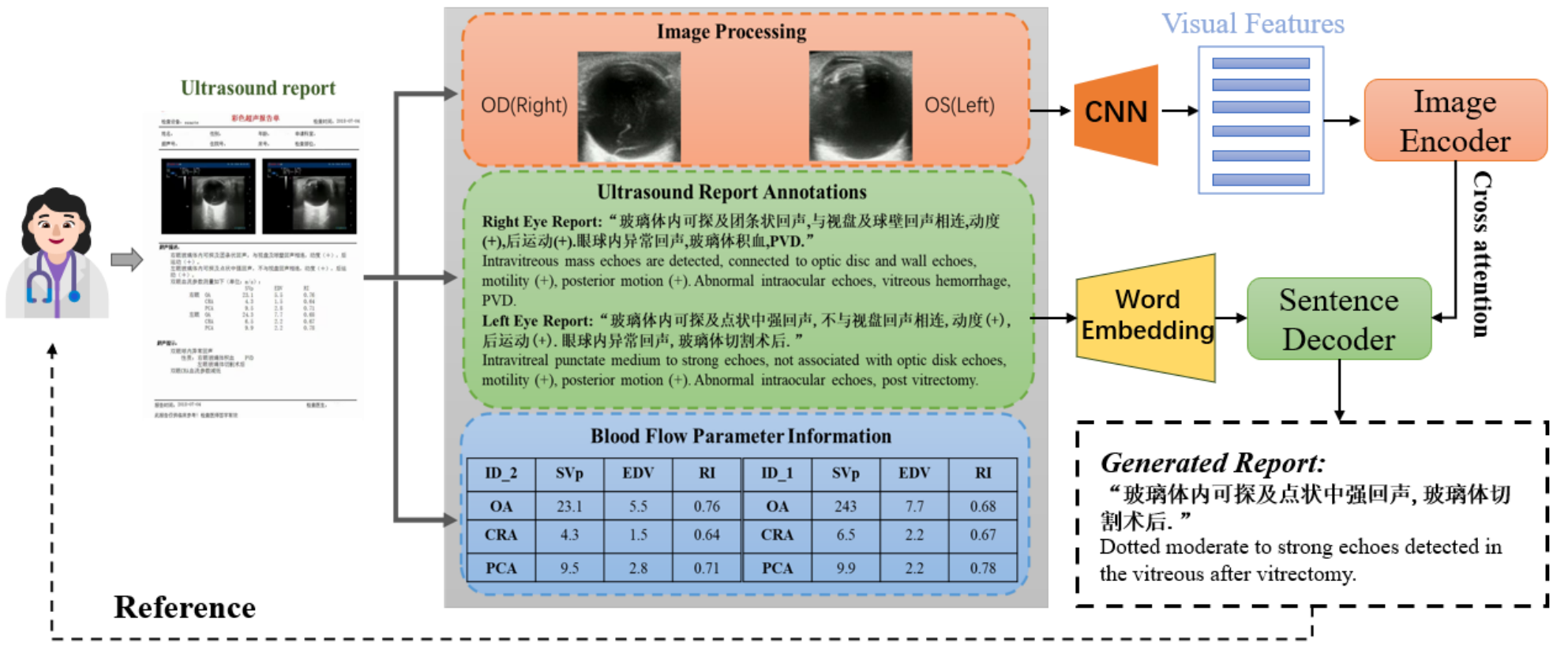} 
	\caption{The framework for dataset constrcution and cross-modal generation} 
	\label{fig1} 
\end{figure}

\subsection{Ultrasound Image Cropping}
A comprehensive retrieval was conducted on all patients treated at Shenyang Eye Hospital in China in 2018. The ocular ultrasound image was taken by the hospital's \textbf{esaote} device. The association between ultrasound image and diagnostic report was established by medical record. Each report of one patient was linked to one or multiple images shown his pathologic status.

Three principles should be followed during the screening of ultrasound image.

(1) The lesions or abnormalities are primarily conscentrated in the eyeball region. Manual cropping of ultrasound image is necessary to cut the irrelevant regions outside the eyeball, in order to reduce the interference on deep learning recognition and diagnosis.

(2) Many images includes color-coded blood flow information and boundary lines during the ultrasound examination. It also introduces interference for image recognition and diagnosis. To ensure a clean dataset, we specifically select these images without any interference from blood flow information.

(3) Some ultrasound images may have off-center eye information due to the variability in eye positioning, and incomplete eyeball is obtained after cropping. So these image should be excluded from dateset. Additionally, the images without corresponding textual reports are also omitted.

\begin{figure}[htbp] 
	\centering 
	\includegraphics[width=0.9\textwidth]{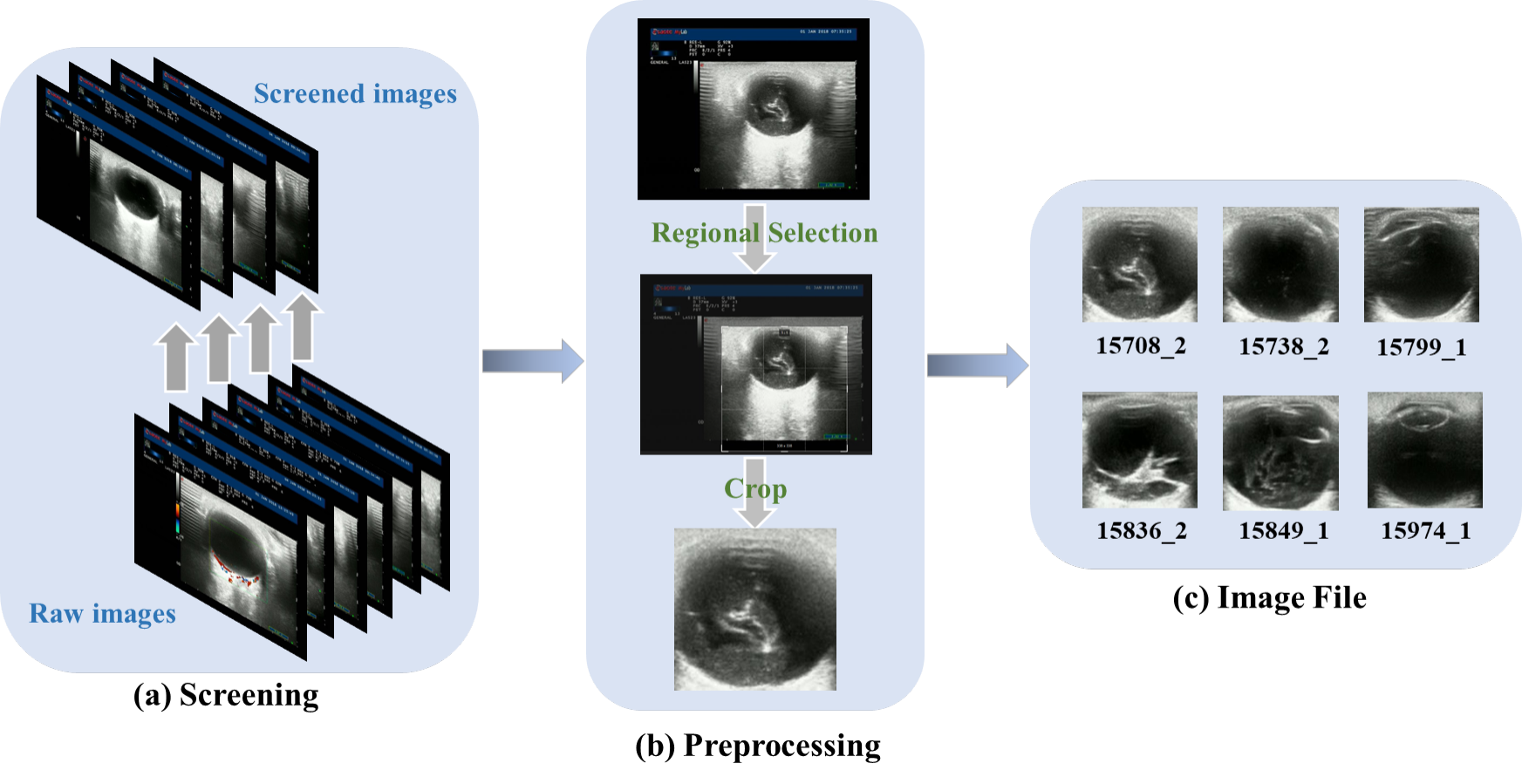} 
	\caption{An example of complete image processing. (a) Screening (b) Regional selection and manual cropping. (c) Storage. } 
	\label{fig2} 
\end{figure}

The image screening and cropping consists of three steps as shown in Fig. 2. Firstly, ultrasound images are selected based on Principle (1-3), and those with color flow interference are removed. Secondly, region selection is performed on the screened images according to the lesion information, then is cropped into square patches. The size of original ultrasound image is 640×480 pixels. Due to the different size and position of the eyeball in an original image,  it is not possible to perform an automatic batch cropping. The image dimension of manual cropped patch ranges from 161×161 to 257×257 pixels, and all patches are saved in JPEG format.

\subsection{Report Preprocessing}
The text dataset included 4858 examination reports. A complete diagnostic report usually contains personal information (name, gender, age), clinical information (ultrasound number, admission number, bed number), and the diagnostic text section. It includes ultrasound description (location and condition of the disease) and ultrasound finding (confirmation of the nature of the disease and diagnosis). An example of clinical report is shown in Figure 3, in which the patient's personal identifiable information is removed.  The following principles are considered during the text extraction of diagnositic report.

(1)	The ultrasound description and its corresponding finding text in the report are integrated. This resulted in a comprehensive textual representation that included detailed disease description and diagnostic finding.

(2)	Some diagnostic texts provided the size parameter to describe the lesion, such as "measuring approximately 1.83×1.48×2.79mm". Considering the image did not provide the specific size information, we delete this parameter and replace it by a simplified description, such as "a hypoechoic cystic-solid lesion within the nasal quadrant."

(3)	Typically, many diagnostic reports include separate diagnoses for the left eye and the right one. Therefore, we process the reports independently for each eye which forms individual case.

(4)	The ultrasound findings did not specify diagnoses for the left and right eyes due to varied writing habits. The report simply mentioned "posterior scleral staphyloma" instead of indicating "left (right) eye with posterior scleral staphyloma". After the confirmation of ophthalmologists, it implies the presence of the disease in both eyes.

We use OCR text recognition technology to identify and extract text from the clinical reports. The ultrasound description and diagnostic section are manually integrated, and the relevant disease description and diagnostic result are retained. Non-relevant text is removed, and the results are differentiated for the left and right eye. An example of the extracted result is shown in Figure 3.

\begin{figure}[htbp] 
	\centering 
	\includegraphics[width=1\textwidth]{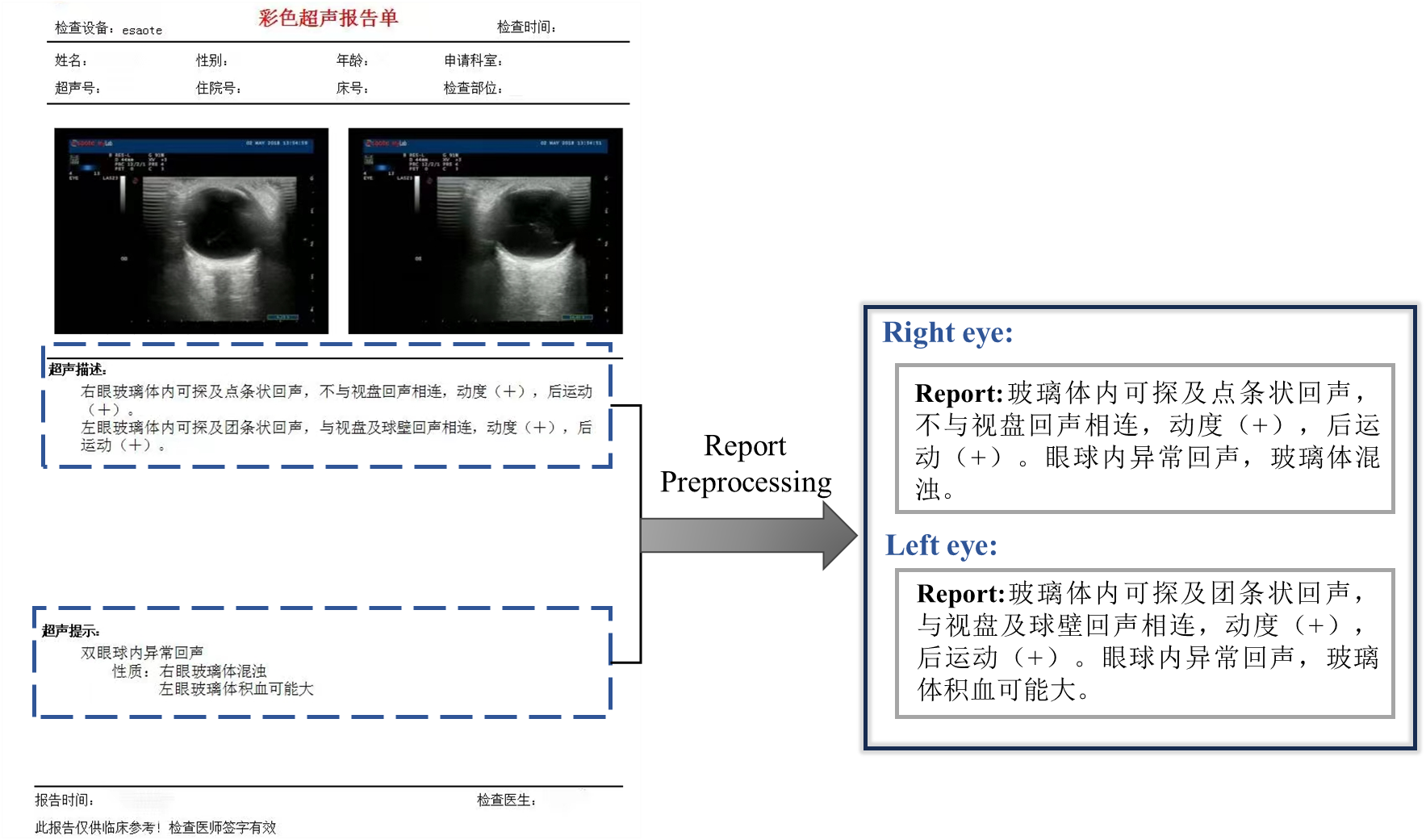} 
	\caption{Example of report preprocessing.} 
	\label{fig3} 
\end{figure}

\subsection{Blood Flow Parameter Recording}
Color Doppler Flow Imaging (CDFI) is primarily used to observe the blood flow parameters when it is applied to ophthalmic testing. It includes the blood flow distribution, flow direction, flow properties, flow velocity, and spectral characteristics at the specific location. The involved detection indices are the end-diastolic velocity (EDV), resistance index (RI), and systolic peak velocity (SVp) of ophthalmic artery (OA), central retinal artery (CRA), and posterior ciliary artery (PCA). We independently collected the CDFI ultrasound images which contain the blood flow measurement. As shown in Figure 4, the measurements of blood flow indices are shown on the left side of the CDFI images. These indices were measured by doctors during the clinical diagnosis process. We extract and compile the information separately. The blood flow indices are extracted from three ultrasound images, and gathered into a $3\times 3$ matrix with nine parameter values. For some cases where the parameter results are shown in the diagnostic report, we read them directly from the report.

\begin{figure}[htbp] 
	\centering 
	\includegraphics[width=0.6\textwidth]{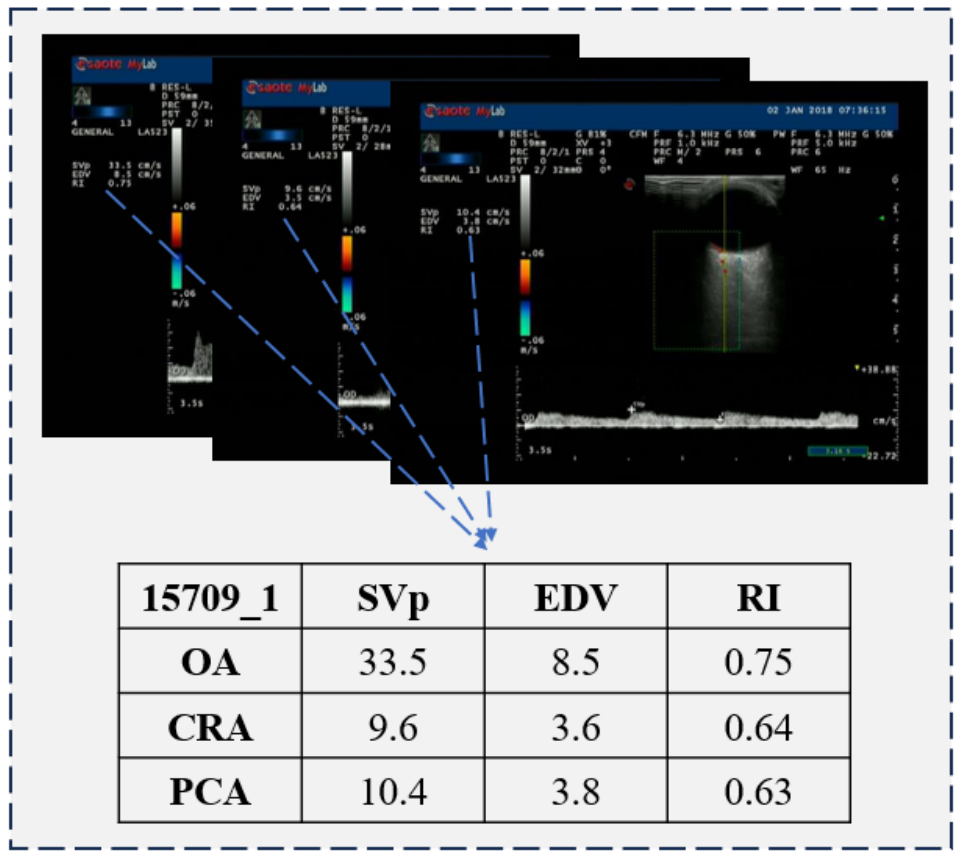} 
	\caption{Example of blood flow information extraction.} 
	\label{fig4} 
\end{figure}

\subsection{Dataset Componuding }
The above three parts of data information are summarized into dataset. First, the cropped image is named according to the corresponding ultrasound number in the clinical report, as shown in Figure 5(a). The left eye image will be named "Ultrasound Number\_1" and the right eye will be named "Ultrasound Number\_2". The subscript "1" or "2" represents the left and right eyes, respectively. If there are multiple images for the same eye, we use the second subscript to discribe the image number. For example, "Ultrasound Number\_11" (left eye) or "Ultrasound Number\_22" (right eye) represent the second image of the left or right eye, respectively. Similarly, the third image of the same eye is named as "Ultrasound Number\_12" (left eye) or "Ultrasound Number\_23" (right eye). The purpose of naming based on the ultrasound number is to facilitate the correspondence with the report information. All the cropped and named images are collected in the same folder.

Next all the image names in the floder are batch extracted and exported to a spreadsheet. This spreadsheet contains the ultrasound number and the corresponding eye designation. The ultrasound report is matched with the image based on its ultrasound number, as well as the image path information. In order to make the data uniformly distributed, we randomly disrupt the data and divid them into training, testing, and validation sets. Finally, the data are converted to JSON format for easy retrieval in the subsequent experiments. The aggregated blood flow parameter matrix is compiled in a table, which also is matched based on the ultrasound number.

In summary, a complete data consists of the image ID (i.e., image ultrasound number), as well as its corresponding report, image path, split set and blood flow information. The report and image are connected by the image ID and its file path. An example of complete data with image and its corresponding report is shown in Figure 5.

\begin{figure}[htbp] 
	\centering 
	\includegraphics[width=0.8\textwidth]{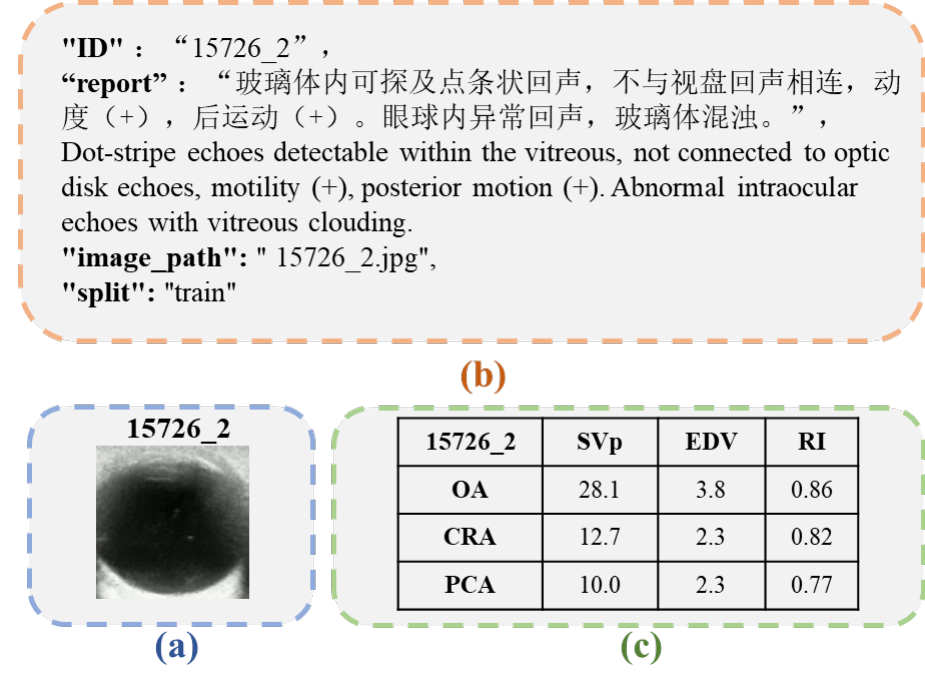} 
	\caption{An example of complete data. (a) cropped image. (b) JSON-formatted content after extracting the text report. (c)  extracted blood flow information.} 
	\label{fig5} 
\end{figure}

\subsection{Statistical analysis}
A comprehensive statistical analysis is conducted to investigate the distribution of disease categories and textual content in the proposed dataset. There is a variety of diseases shown in this dataset. The 15 most common diseases are shown in Figure 6, in which the diseases less than 5 occurrences are not considered. The most common disease are "vitreous opacity," "vitreous hemorrhage," "PVD(Posterior Vitreous Detachment)" and "mild vitreous opacity". The number of each disease is calculated based on the frequency of the corresponding keyword in the dataset.

\begin{figure}[htbp] 
	\centering 
	\includegraphics[width=1\textwidth]{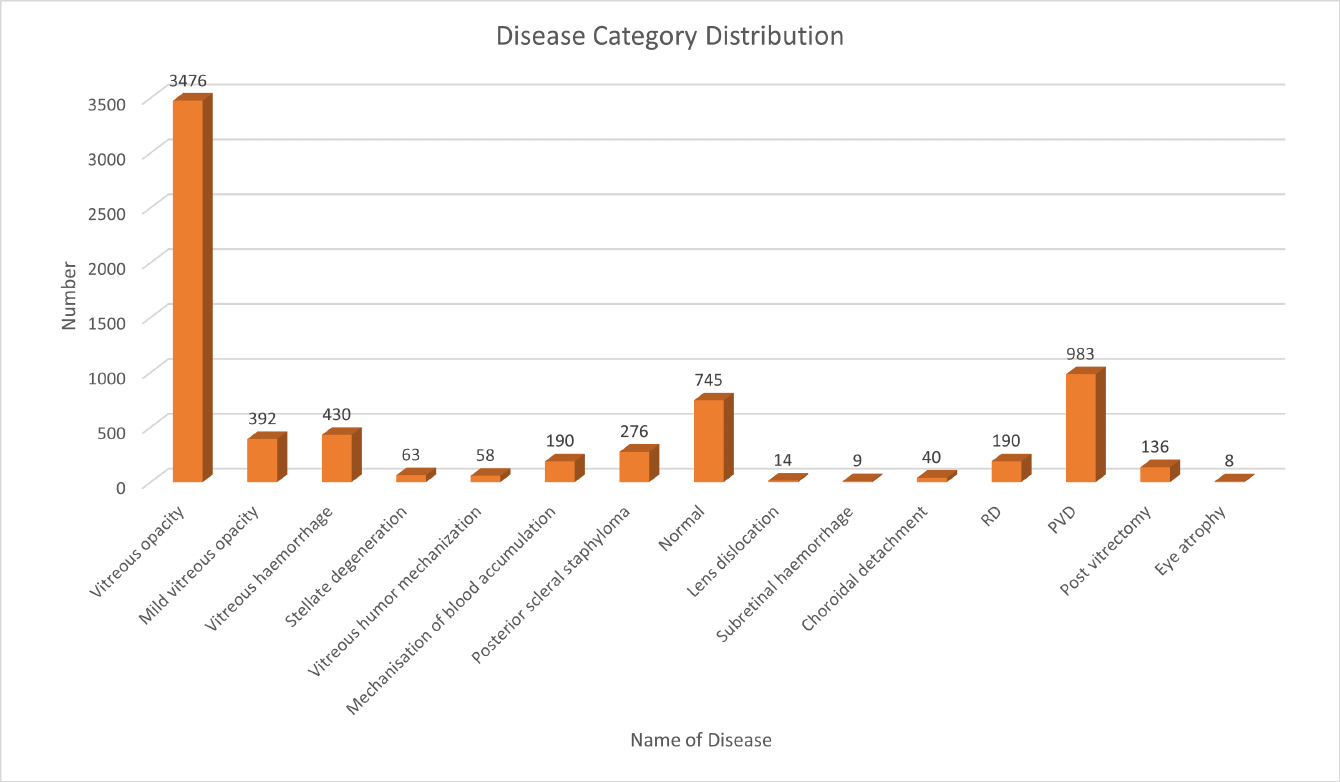} 
	\caption{Disease categories and percentages.} 
	\label{fig6} 
\end{figure}

It is shown that 735 nomal cases are found at 15.3\% of the total dataset, and no significant ophthalmic disease is found in these patients. The symptom "vitreous opacity" is 3476 cases at 71.6\% of the total dataset, and it is the most common eye diseases. In contrast, "eye atrophy" case is rare with only 8 cases at 0.016\%. The number of left and right eye cases in the dataset are statistically 2,394 and 2,465, respectively. The right-eye samples are slightly more than the left-eye samples. The distribution of left and right eyes is unbalanced, since different patients underwent one or multiple ultrasound scanning at different eyes, angle and times. In addition, other necessary index of the dataset are analyzed, including the number of reports and patients, the length of sentences in the reports. The detailed results are shown in Table 2. It is note that all sentences are extracted from diagnostic reports written by physicians, and each report consists of a series of sentences. There is a significant difference in the sentence length of the report. For example, the maximum or minimum tokens in a single report is 114 and 9, respectively. Usually, the report of normal case is described simplely as "No abnormality was observed in the specific area". The report of complex case combines the description of multiple diseases at specical locations, and its sentence length is far longer than that of the normal case.

\begin{table}[htbp]
	\begin{center}
		\caption{Descriptive statistics of report sentences. }
		\begin{tabular}{cc}
			\hline \textbf{Parameter} & \textbf{Value} \\
			\hline Total tokens & 252676 \\
			Reports & 4858 \\
			Patients & 2417 \\
			Average (min-max) number of tokens per report & 62 \\
			Maximum tokens in a single report & 114 \\
			Minimum tokens in a single report & 9 \\
			Right eyes & 2393 \\
			Left eyes & 2465 \\
			\hline
		\end{tabular}
	\end{center}
\end{table}

\section{Medical Report Generation}
\subsection{Methodology}

In order to evaluate the usability of the proposed dataset, cross-modal memory network (CMN) is adopted for the automatic generation of medical report. It introduces a CMN module to enhance the encoder-decoder framework during the chest radiology report generation \cite{ref26}. In detail, it designs a shared memory to capture the interactive information between image and text. Then the interacted or alignment infromation are fed into the encoder and decoder of transformer, respectively, which can facilitate the cross-modal interaction and generation. In order to apply CMN for Chinese report learning, we additionally used the jieba word-splitting tool to split the text to get some medical terms about ophthalmic ultrasound images. This will be more suitable for Chinese medical report generation task.

A schematic framework is given to illustrate the automatic generation of the ophthalmic ultrasound report based on CMN learning model, as shown in Figure 7. The whole process consists of four modules: visual feature extraction, word embedding, cross-modal memory network and report generation. The visual feature extraction module utilizes ResNet to extract the patch features from ultrasound image. The word embedding module transforms text into vector form. The cross-modal memory network projects the image features and text vectors into the same space and facilitates interaction using a memory matrix, in order to align the information between different modals, such as image and text. Finally, the interacted image and text information are fed into the  encoder and decoder of transformer, respectively, and the report is generated.

\begin{figure}[htbp] 
	\centering 
	\includegraphics[width=1\textwidth]{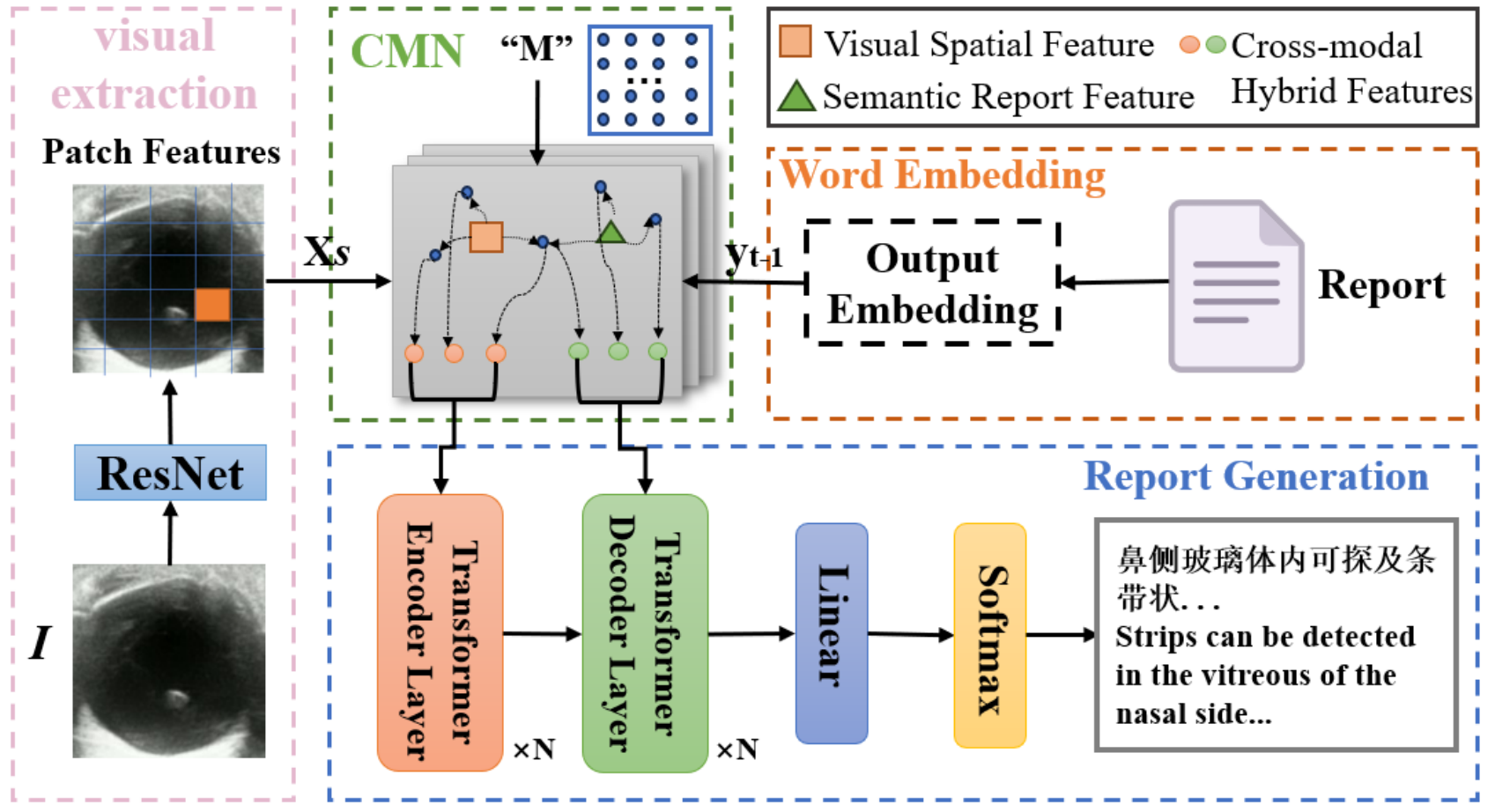} 
	\caption{A framework diagram for cross-modal medical report generation.} 
	\label{fig7} 
\end{figure}

We define the generation of ultrasound report as an image-to-text generation task. The goal is to predict the diagnostic report $ \mathbf{R} $ for each ultrasound image $ \mathbf{I} $. Then the image can be used as the source sequence and the text report as the target sequence.

\textbf{Visual feature extraction:} First, a visual extractor is used to extract the visual features from the ultrasound image $ I $, which is denoted as $\mathbf{X}=\left\{\mathbf{x}_1, \mathbf{x}_2, \ldots, \mathbf{x}_S\right\}, \mathbf{x}_s \in \mathbb{R}^d$. Here $ \mathbf{x}_s $ are the patch features of the image and $ d $ is the size of the feature vector. The extraction process is expressed as:
\begin{equation}
	\left\{\mathbf{x}_1, \mathbf{x}_2, \ldots, \mathbf{x}_S\right\}=f_v(\mathbf{I})
\end{equation}
where the visual extractor $f_v(\cdot)$ is finished by ResNet network \cite{ref9} in this experiment.  

\textbf{Word embedding:} The text sequence obtained by word embedding is denoted as $ \mathbf{Y}=\left\{\mathbf{y}_1, \mathbf{y}_2, \ldots,\mathbf{y}_{t-1}\right\}$. The process can be expressed as:
\begin{equation}
	\left\{\mathbf{y}_1, \mathbf{y}_2, \ldots,\mathbf{y}_{t-1}\right\}=f_t(\mathbf{R})
\end{equation}
where $f_t(\cdot)$ is the text feature extractor. Here we used the same embedding module of report generation Transformer during the model training and learning. The generated output tokens are represented as $ \left\{\mathbf{y}_1, \mathbf{y}_2, \ldots, \mathbf{y}_t, \ldots, \right.$ \\$\left. \mathbf{y}_T\right\} , \mathbf{y}_t \in \mathbb{V}$, where $ \mathbb{V} $ is all the possible tokens and $T$ is the length of the report.

\textbf{Cross modal network:} CMN is used to solve the information alignment problem from different modalities. It introduces a memory matrix $\mathbf{M}=\left\{\mathbf{m}_1, \mathbf{m}_2, \ldots, \mathbf{m}_i, \ldots, \mathbf{m}_{\mathcal{N}}\right\}$ to deal with the information interaction between two modalities including image and text. Here $ \mathcal{N} $ denotes the number of memory vectors. Specifically, CMN consists of two subprocesses, querying and responding. In the query subprocess, the image and text features are first projected into the same representation space, and the most relevant memory vectors about the image and text are queried. The responding subprocess is to weight the queried memory vectors of image and text, respectively. Finally, the obtained memory responses are fed into the Transformer to generate the corresponding reports. The CMN learning can be represented as:
\begin{equation}
	\left\{\mathbf{r}_{\mathbf{x}_1}, \mathbf{r}_{\mathbf{x}_2}, \ldots, \mathbf{r}_{\mathbf{x}_S}\right\}=CMN \left(\mathbf{x}_1, \mathbf{x}_2, \ldots, \mathbf{x}_S\right)
\end{equation}
\begin{equation}
	\left\{\mathbf{r}_{\mathbf{y}_1}, \mathbf{r}_{\mathbf{y}_2}, \ldots, \mathbf{r}_{\mathbf{y}_{t-1}}\right\}=CMN \left(\mathbf{y}_1, \mathbf{y}_2, \ldots, \mathbf{y}_{t-1}\right)
\end{equation}
where $\mathbf{r}_{\mathbf{x}_i}$ and $\mathbf{r}_{\mathbf{y}_j}$, $i=1,\ldots,S$, $j=1,\ldots,t-1$ are the memory responses of visual and textual features. The detailed description about CMN querying and responding can be found in \cite{ref26}.

\textbf{Report generation:} The report generation is finished in Transformer architecture. The CMN image information $  \left(\mathbf{r}_{\mathbf{x}_1}, \mathbf{r}_{\mathbf{x}_2}, \ldots, \mathbf{r}_{\mathbf{x}_S}\right)  $ is input to the encoder of Transformer to get the intermediate state $ \left\{\mathbf{w}_1, \mathbf{w}_2, \ldots, \mathbf{w}_S\right\} $, which is expressed as:
\begin{equation}
	\left\{\mathbf{w}_1, \mathbf{w}_2, \ldots, \mathbf{w}_S\right\}=f_e\left(\mathbf{r}_{\mathbf{x}_1}, \mathbf{r}_{\mathbf{x}_2}, \ldots, \mathbf{r}_{\mathbf{x}_S}\right)
\end{equation}
where $f_e(\cdot)$ refers the encoder. Then along with the memory response $ \left\{\mathbf{r}_{\mathbf{y}_1}, \mathbf{r}_{\mathbf{y}_2}, \right.$ \\$\left. \ldots, \mathbf{r}_{\mathbf{y}_{t-1}}\right\} $, the obtained intermediate states $ \left\{\mathbf{w}_1, \mathbf{w}_2, \ldots, \mathbf{w}_S\right\} $ are sent to the decoder  to generate the current output $ y_{t} $.
\begin{equation}
	\mathbf{y}_t=f_d\left(\mathbf{w}_1, \mathbf{w}_2, \ldots, \mathbf{w}_S, \mathbf{r}_{\mathbf{y}_1}, \mathbf{r}_{\mathbf{y}_2}, \ldots, \mathbf{r}_{\mathbf{y}_{t-1}}\right)
\end{equation}
where $f_d(\cdot)$ refers the decoder.

\textbf{Loss function:} It is found that the entire report generation is recursive with the visual features $\mathbf{X}=\left\{\mathbf{x}_1, \mathbf{x}_2, \ldots, \mathbf{x}_S\right\}$  from the ultrasound image $  \mathbf{I} $ as input, and  the report target sequence $ \mathbf{Y} = \left\{\mathbf{y}_1, \mathbf{y}_2, \ldots,\mathbf{y}_t, \ldots, \mathbf{y}_T\right\}$ as output. It can be formalized based on the recursive chain rule as follows:
\begin{equation}
	p(\mathbf{Y} \mid \mathbf{I} )=\prod_{t=1}^T p\left(\mathbf{y}_t \mid \mathbf{y}_1, \ldots, \mathbf{y}_{t-1}, \mathbf{I}\right)
\end{equation}
where $ p(\mathbf{Y} \mid  \mathbf{I} ) $ is the probability of generating the target sequence $ Y $ for the given input $ I $. Then the model training is to maximize the conditional log-likelihood of $ p(\mathbf{Y} \mid \mathbf{I} ) $ :
\begin{equation}
	\theta^*=\underset{\theta}{\arg \max } \sum_{t=1}^T \log p\left(\mathbf{y}_t \mid \mathbf{y}_1, \ldots, \mathbf{y}_{t-1}, \mathbf{I} ; \theta\right)
\end{equation}
where $ \theta $ are the model parameters. The goal is to find the best parameter value $ \theta^* $ to maximize the probability of generating the target sequence $ \mathbf{Y} $ for the given input $ \mathbf{I} $.	

\subsection{ Experiments and Discussions}
\subsubsection{Dataset and Evaluation Metrics}
We randomly separate the ultrasound dataset into three parts for training, validation and test according to the ratio of 75:10:15. The model is trained and validated using the training and validation sets. The test set is adopted for performance evaluation, which is not available during the training procedure. The detailed data split is shown in Table 3.
\begin{table}[htbp]
	\begin{center}
		\caption{Detailed information of data split.}
		\setlength{\tabcolsep}{2mm}{
		\begin{tabular}{ccccc}
			\hline  Dataset  &Training & Validation&Test&Total \\
			\hline
			Images & 3644 & 485 & 729 & 4858 \\
			Reports & 3644 & 485 & 729 & 4858 \\
			\hline
		\end{tabular}}
	\end{center}
\end{table}

In this paper, four Natural Language Generation (NLG) metrics are used to evaluate the model quality of ultrasound reports  generation. They are  BLEU \cite{ref27}, METEOR \cite{ref28}, CIDER \cite{ref29} and ROUGE-L \cite{ref30}. In particular, BLEU measures the similarity between generated and truth text by calculating the overlap of word n-grams. METEOR takes into account the superficial and semantic similarities between the generated text and the truth text. It also has built-in mechanisms for handling synonyms and paraphrases. ROUGE-L is a kind of evaluation based on the precision and recall under the longest common subsequence $L$. The semantic content and coherence of the generated text are also considered. In contrast, CIDER is based on the cosine similarity between word embedding concurrently considering both single-word phrases and multi-word phrases. It is a metric used to evaluate image description and text generation tasks. Here we particularly use CIDER to evaluate the capture of important information during the generated task.

\subsubsection{Experiment Details}
We use ResNet-101 \cite{ref9} backbone pre-trained on ImageNet \cite{ref32} as the visual extractor to extract the patch visual features. The dimension of each feature is 2,048. 60 epochs are trained and the batch size is 16. All the Ultrasound images are firstly resized to 224 × 224. Adam optimizer is adopted to train the model. The learning rates of the visual extractor and other parameters are set to 5e-5 and 1e-4, respectively. The whole training framework is implemented with a PyTorch 1.7.1 library based on Python3.8 and NVIDIA GeForce RTX 3090 32GB GPU. For the encoder-decoder backbone in report generation Transformer, its structure includes 3 layers and 8 attention heads, with 512 dimensions of hidden states. All weights in the model are randomly initialized. The beam size is 3 to balance the effectiveness and efficiency of all models during the report generation. The model who achieves the best BLEU-4 score on the validation set serves as the best trained model.

The cleaning for text data is necessary in order to improve the quality of report generation. The length of the sequence in a single report is basically less than 110 characters. Therefore, the maximum sequence length is defined as 115 during the training process. For these sentences that are not long enough, the special character [unk] is used to fill them. The cut-off threshold for words is 3. It means the word with fewer occurrences than the threshold is filtered out and instead marked with [unk]. In addition, the disease abbreviation in the original report is replaced with its corresponding Chinese name. For example, "PVD" is replaced with "posterior vitreous detachment".

To validate the generalizability of proposed dataset, we also conducts other generation model R2Gen \cite{ref23} as the main baselines in our experiments. R2Gen model uses a relational memory (RM) to record the previous generation process and combines memory-driven conditional layer normalization (MLCN) in the Transformer decoder. The ablation experiments for two models (R2Gen and CMN) on the proposed dataset are designed, respectively.

BASE: This is an original Transformer with 3 layers, 8 heads and 512 hidden units, with no other extensions or modifications.

BASE+RM: The RM module is connected directly to the converter output before the softmax of each time step, but not integrated into the Transformer decoder.

BASE+R2Gen: The MLCN and RM modules are combined and integrated in the Transformer decoder to facilitate the decoding process.

BASE+MEM: Two memory networks for image and text are introduced into the BASE Transformer, but there is no cross-modal information interaction.

BASE+CMN: A shared memory network CMN is introduced to facilitate information interaction between two modalities, image and text.

\subsubsection{Results and Analyses}
The four NLG metrics obtained from the ablation experiments are shown in Table 4. It is shown that various models achieve good report generation performance on the proposed dataset. This indicates that the proposed dataset can be serviced as a standard verification for different automatic medical report generation methods. The structure of proposed ophthalmic dataset also gives a guide for other image-text medical report dataset.

\begin{table*}[t!]
	\begin{center}
		\caption{The NLG performance metrics on the test set.}
		\label{t7}
		\setlength{\tabcolsep}{0.9mm}{
			\begin{tabular}{cccccccc}
				\cline{1-8}
				\multirow{1}{*}{} &  \multicolumn{7}{c}{\textbf{NLG METRICS}} \\ 
				\cmidrule(r){2-8} 
				\textbf{MODEL} & \textbf{BL\_1}& \textbf{BL\_2}& \textbf{BL\_3}& \textbf{BL\_4}& \textbf{METEOR}& \textbf{ROUGE-L}& \textbf{CIDER}\\ 
				\cline{1-8}
				 BASE & 0.382 & 0.309 & 0.154 &0.126  & 0.368 & 0.339 & 0.499 \\
				 BASE+RM & 0.470 & 0.364 & 0.219 & 0.165 & 0.405 & 0.371 & 0.550 \\
				 BASE+R2Gen & 0.568 & 0.457 & 0.344 & 0.248 & 0.475 & 0.495 & 0.834 \\
				 BASE+MEM & 0.487 & 0.390 & 0.294 & 0.184 & 0.457 & 0.365 & 0.637 \\
				 BASE+CMN & 0.589 & 0.459 & 0.396 & 0.270 & 0.493 & 0.511 & 0.951 \\
				\cline{1-8}
		\end{tabular}}
	\end{center}
\end{table*}

Besides the NLG metrics, another important criterion for evaluating the generation model performance is the length of generated report. The closer the length of the generated report is to the ground-truth report length, the better the generation model performance. We divide the generated reports into 12 groups based on their report lengths (the word range is [0, 120]) with intervals of 10. Figure 8 shows the comparison results of the report length distribution in each intervals generated by two models R2Gen and CMN. The majority length of ground-truth reports in the test set concentrates around 50-60 words, with approximately 350 reports falling in this range. It can be found that the length of generated report both closely match with that of the ground-truth report in intervals [10,30) and [70,120]. The report generated by CMN model is closer to the ground-truth reports than the R2Gen model in the interval [30,70). Overall, the proposed dataset is applicable to the cross-modal medical report generation task and provides a good validation for the medical report generation algorithm.

\begin{figure}[htbp] 
	\centering 
	\includegraphics[width=0.94\textwidth]{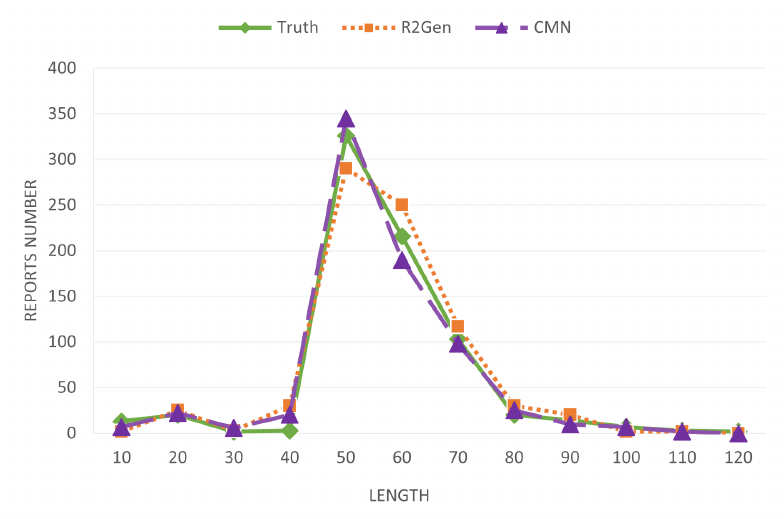} 
	\caption{Comparison of the report length.} 
	\label{fig8} 
\end{figure}

We use heat map to visualize the mapping relationship between image and text. Figure 9 shows the visualization results of specific regions in the ocular ultrasound image (represented by color weights) with the reports generated by CMN model. The color plots represent the weight values from low to high in the range [0, 1]. The professional ophthalmologist gives the interpretation of the visualization results. They point out the areas in the images where lesions are present. Based on the descriptions provided by the doctors and the visualized results, the model demonstrates a good ability to focus attention on the corresponding lesion areas. For example, the model accurately identifies the location of lesions such as "vitreous hemorrhage", "post-vitrectomy" and "vitreous opacities" in the images.

In Figure 9, the red font represents that the model failed to predict or predicted incorrectly during the report generation process. The blue font represents the additional portion that the model gives more prediction information than the true report. In the first example, the model does not accurately distinguish between "vitreous hemorrhage" and "vitreous mechanized accumulation of blood". They are two diseases that are very easily confused both in lesion image and in textual description. In the second and third examples, the CMN model accurately predicts the report reality. The interesting area in visualization image focuses on the lesion area, which implies that the model is able to align the image and text information well. In the fourth example, the model not only accurately predicts the disease information, but also additionally generates a phrase "abnormal intraocular echoes". It shows the strong learning ability of the CMN model. In addition to the successful predictions, there are also some interferences present in the dataset. For instance, in the fifth example, the "posterior scleral staphyloma" is located at the lower edge of the eye image, which is difficult to recognize by the CMN model. In addition, due to the visual similarity between the surrounding non-ocular regions and the lesion region, it causes an additional interference which affects the accuracy identification and report generation. In general, the result of  visualization and the corresponding generated report indicates that the proposed dataset is suitable for medical report generation.

\begin{figure}[htbp] 
	\centering 
	\includegraphics[width=1\textwidth]{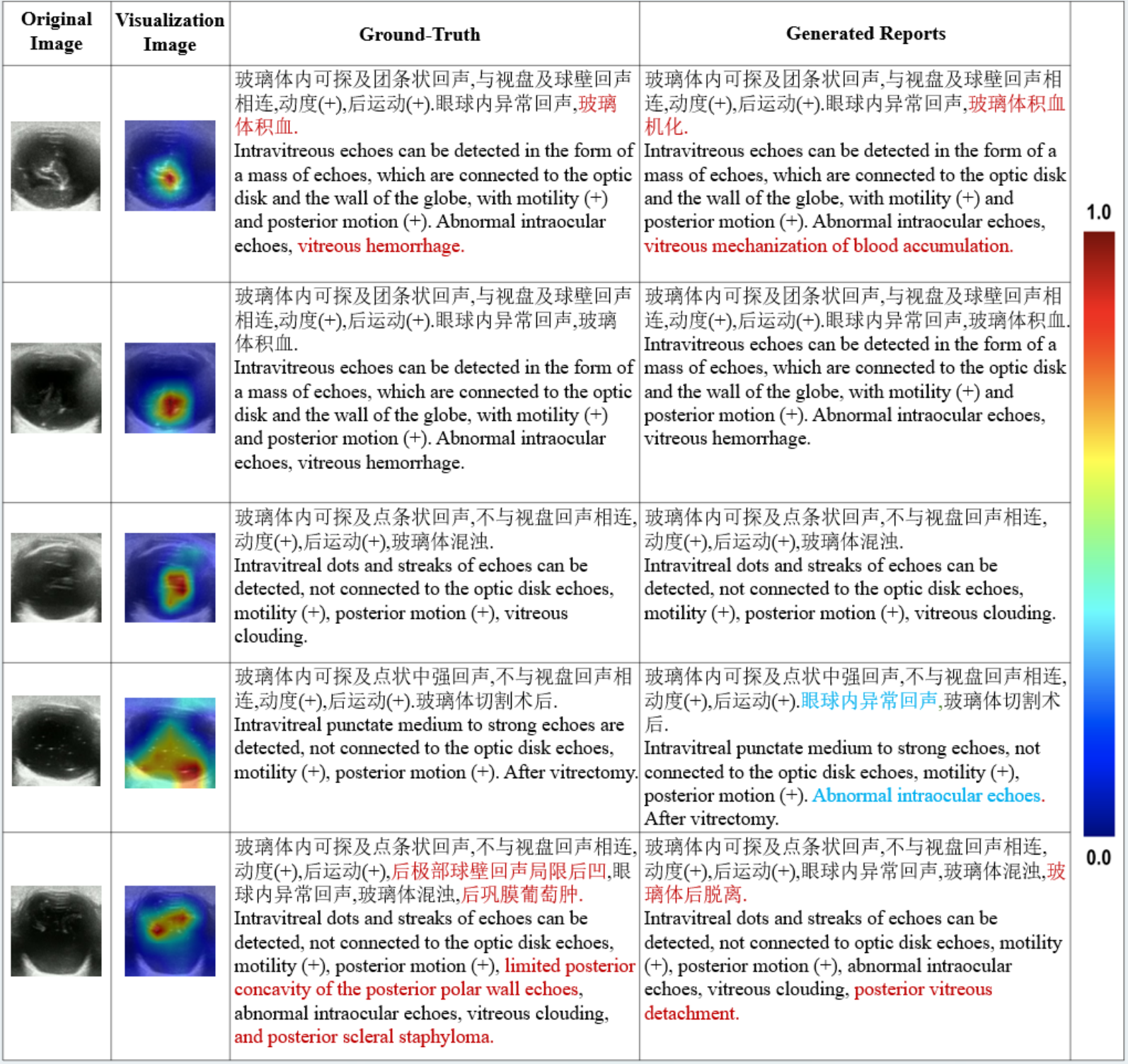} 
	\caption{Visualizations of image-text attention mappings of generated report from CMN model.} 
	\label{fig9} 
\end{figure}

\section{Conclusion and Limitations}
This paper presents a labeled ophthalmic dataset for the precise analysis and the automated exploration of medical image along with its associated report. The dataset contain 4858 Chinese reports and the corresponding 4858 eye ultrasound images, as well as the information of blood flow parameters measured in clinical practice. To the best of our knowledge, it is the only ophthalmic dataset that contains the three modal information simultaneously. The proposed dataset has also been used to evaluate the cross-modal medical report generation models including the R2Gen and CMN models. The accuracy report generation and its corresponding interesting disease areas are also visualized based on CMN model. We hope that this dataset can contribute to the development of automated diagnostic learning algorithm for ophthalmic domain and reduce the stress of ophthalmologist in their clinical work.

We also notice that there are several limitations of this study. First, all these data are collected from only one medical center and may not be generalizable. Second, there are still some rare disease variants not collected in the dataset. Third, there is a data bias in the distribution of diseases because the data are collected in a real clinical process. In the future, we will continue to expand the volume of dataset to minimize the data bias as much as possible.




\section*{Acknowledgments}
This research is funded by the National Natural Science Foundation of China (62373005, 62273007).

\bibliographystyle{elsarticle-num}
\bibliography{refs}

\end{document}